# Deep Diabetologist: Learning to Prescribe Hypoglycemia Medications with Hierarchical Recurrent Neural Networks

Jing Mei[a], Shiwan Zhao[a], Feng Jin[a], Eryu Xia[a], Haifeng Liu[a], Xiang Li[a]

[a] *IBM Research China, Beijing, China*

**Abstract**

*In healthcare, applying deep learning models to electronic health records (EHRs) has drawn considerable attention. EHR data consist of a sequence of medical visits, i.e. a multivariate time series of diagnosis, medications, physical examinations, lab tests, etc. This sequential nature makes EHR well matching the power of Recurrent Neural Network (RNN). In this paper, we propose "Deep Diabetologist" – using RNNs for EHR sequential data modeling, to provide the personalized hypoglycemia medication prediction for diabetic patients. Particularly, we develop a hierarchical RNN to capture the heterogeneous sequential information in the EHR data. Our experimental results demonstrate the improved performance, compared with a baseline classifier using logistic regression. Moreover, hierarchical RNN models outperform basic ones, providing deeper data insights for clinical decision support.*

## Introduction

From a global report on diabetes [1], over 422 million adults have been diagnosed with diabetes, by 2014. In China, there have been 114 million adults suffering from diabetes, by 2013 [2]. Roughly estimated, in China, a typical endocrinology visit in the outpatient clinic has only 5-10 minutes, and often, a diabetologist may see up to 50 patients in a single morning [3].

Diabetes is a chronic disease, and hypoglycemia medications need to be timely adapted according to the clinical conditions. Most Chinese diabetic patients go to hospitals, regularly (generally 2-4 weeks) for prescribing the hypoglycemia medications. Facing such a large scale of diabetic patients, but with a very limited coverage of diabetologists in China, is there any way to learn to prescribe hypoglycemia medications by deep learning from data (so as to provide clinical decision support for diabetologists)?

To answer this question, we need two key points. One is what's the data, and the other is what's the learning framework. Luckily, with the development of regional health information systems in China, Electronic Health Record (EHR) repositories have been growing up with a great amount of longitudinal patient visits [4]. Taking a level 2 city in China as an example, there have been 3.6 million patients documented in her EHR repository, having a time window from 2006 to present (over 10 years). Undoubtedly, advanced deep learning models are expected to shed light on the meaningful use of EHR.

Considering that EHR data consist of multivariate time series of observations (such as diagnosis, medications, physical examinations and lab tests), the well-matched deep learning framework is Recurrent Neural Network (RNN). RNN captures the sequential nature of EHR data, and recently there have been at least three remarkable representatives on applying RNNs to EHR data analysis. In [5], Lipton et al. used Long Short-Term Memory (LSTM, one of RNN implementations) to predict diagnosis codes in the pediatric intensive care unit. Their network input was a vector of 13 physical examinations and lab tests, for training a LSTM model to classify 128 diagnosis codes. In [6], Choi et al. used Gated Recurrent Units (GRU, one of RNN implementations) to predict both the diagnosis codes and the medication codes, as well as the time duration until next visit, among primary care patients. Their network input was a vector of 38,594 diagnosis and medication codes, for training a GRU model to classify 1,183 diagnosis codes and 595 medication codes, plus a regression of the time duration between visits. A more clinically meaningful exploration is [7], which again used GRU to demonstrate the improved model performance in predicting initial diagnosis of heart failure.

Related works have well proved the RNN power on EHR data analysis, however, we observed that EHR data have not been completely exploited in previous works. For example, the real-valued physical examinations and lab tests were used in [5] but not in [6][7]. Conversely, the discrete diagnosis and medication codes were used in [6][7] but not in [5]. Fortunately, both real-valued measurement variables and discrete code variables have been documented in our EHR repository, and in Figure 1, we illustrate the sample data for a given patient, sorted by date. For instance, on 2015-06-15, this patient had records about fasting plasma glucose (FPG) valued as 8.13 mmol/L, was diagnosed with diabetes (using ICD 10 code of E14), and was prescribed with biguanides (as hypoglycemia drug).

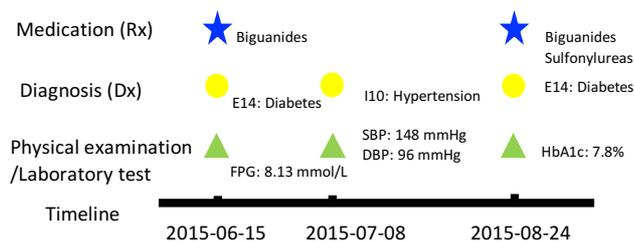

*Figure 1. A sample patient data from EHR repository.*

Observing that time series in EHR data could be layered by different types, e.g. in Figure 1, the bottom layer is for clinical measurements, the middle layer is for diagnoses, and the top layer is for medications, we propose to leverage the hierarchical recurrent neural network (HRNN) framework for a better multi-resolution learning. Intuitively, a series of clinical measurements might be targeted to some diagnosis, while a series of diagnoses (plus a series of clinical measurements) might be targeted to some medication. In literature, HRNN has been proposed and implemented for computer vision [8], natural language processing [9], dialogue response generation [10], etc. However, to the best of our knowledge, HRNN has not been used for EHR data analysis yet.

In this paper, we explore to build RNN in a hierarchical way for hypoglycemia medication prediction, and we name "Deep Diabetologist" which is a cognitive decision advisor to provide clinical decision support for treatment recommendation.

## Methods

Before diving into the details of learning techniques, we first need to do EHR data preprocessing including cleansing and imputation. Then, we will introduce the learning framework of "Deep Diabetologist", including the basic and hierarchical RNN medication prediction models.

### Data Preprocessing

As shown in Figure 1, we access to an EHR repository consisting of several relational tables (including a diagnosis table, a medication table, a physical examination table, a lab test table, etc.) with association of patient IDs and event IDs.

In the medication table, 745 distinct drug brand names with production batches are documented, and for non-commercial purpose, we will map them to the higher level drug classes. By definition of a drug taxonomy (after consulting domain experts), they are mapped into 7 hypoglycemia drug classes: Biguanides, Sulfonylureas, Glinide, Thiazolidinediones (abbr. TZDs), Alpha-glucosidase inhibitors (abbr. AGIs), Dipeptidyl peptidase-4 (abbr. DPP-4), and Insulin.

In the diagnosis table, there are 27,258 distinct ICD 10 codes used (e.g. E11.901 and E11.902). Considering the first 3-digit of an ICD 10 code (e.g. E11) is representative for a specific disease, we extract out 1,419 distinct 3-digit ICD 10 codes. Figure 2 is the scatter plot of these 1,419 codes, where the x-axis represents the codes and the y-axis represents the number of medical visits having the codes. Clearly, it is a long-tail distribution, and we catch the 350 most-frequently used ones as the diagnosis codes in our experiments (since those less used ones might introduce noise in the data).

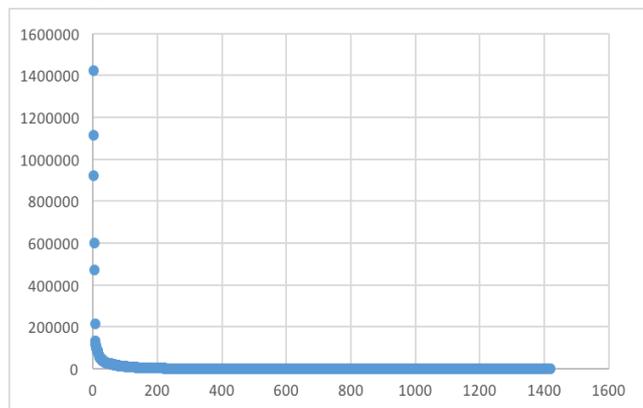

*Figure 2. A scatter plot of ICD codes.*

Also, we conduct data cleansing for the physical examination and lab test tables, leaving only 124 measurement codes. Not surprisingly, this vector of 124 variables has quite a lot missing values, because not every medical visit needs to do all of the measurements. Similar to the work of [5], we do imputation, but differently, we have no such clinically normal values as defined by domain experts for these all 124 variables. Therefore, we first normalize each variable $x$, i.e. rescaled to the standard deviation: $x' = \frac{x-\mu}{\sigma}$ where $\mu$ is the mean of the population, and $\sigma$ is the standard deviation of the population. Next, we impute zero for missing values, since after normalization, zero is for the mean of the population. That is, when a variable value is missing, we assume it as the mean.

Next, we define the cohort as diabetic patients, and the studied cases consist of outpatient hypoglycemia medications. For experiments, 21,796 patients are selected into our cohort, satisfying the following 3 criteria: (1) they are city residents (assuming a city-level EHR repository has continuous records for city residents); (2) they have been diagnosed with diabetes (i.e. documented by ICD 10 codes of E10, E11, E12, E13, E14); (3) they have sequential hypoglycemia medications of length more than 10 from 2006 to present (towards a better RNN sequence learning). Table 1 shows the medical visit numbers of diagnoses, medications and measurements, by counting per day, per patient (and per code), for the cohort of 21,796 patients.

*Table 1. Statistics of medical visits*

|  | Counting per day, per patient, per code | Counting per day, per patient |
|---|---|---|
| Diagnosis | 4,154,756 | 3,371,287 |
| Medication | 805,477 | 620,633 |
| Measurement | 781,501 | 81,294 |

The number of visits with diagnosis codes is over 4 million, and by aligning the different diagnosis codes within one day for one patient, we still get more than 3 million. The number of visits with medications is smaller (because only hypoglycemia medications are counted in), and by average, there are 28.5 medications for a diabetic patient. A weird issue is the quite small size of clinical measurements. By the patient-day alignment, the average number of visits with clinical measurements is only 3.7. After consulting local physicians and patients in the city, we understand this as a common situation: for a chronic disease like diabetes, most patients go to visit physicians, esp. in community hospitals, regularly (generally 2-4 weeks) for prescribing the medications without any clinical measurement. It was said, by policy, the prescription interval at community hospitals has been limited to at most 1 month.

*Table 2. Statistics of using previous medications*

| Previously prescribed == null | 21,796 |
|---|---|
| Previously prescribed <> currently prescribed | 198,153 |
| Previously prescribed == currently prescribed | 400,684 |
| Total | 620,633 |

Then, we have to pay special attention to an important factor in our scenario – the previously prescribed drug classes! As shown in Table 2, for our 620,633 medications (after the patient-day alignment), there are 400,684 cases prescribed as the same as previously, resulting in an accuracy of 400684/620633=0.6456. Row 1 in Table 2 reports 21,796 cases, each of which corresponds to the first prescription of each patient in the cohort.

*Table 3. Statistics of hypoglycemia drug classes*

|  | Number | Ratio |
|---|---|---|
| Biguanides | 141,900 | 0.2286 |
| Sulfonylureas | 88,995 | 0.1434 |
| Glinide | 87,255 | 0.1406 |
| TZDs | 20,317 | 0.0327 |
| AGIs | 163,972 | 0.2642 |
| DPP-4 | 939 | 0.0015 |
| Insulin | 302,099 | 0.4868 |

Finally, Table 3 is presented to show the imbalance of cases being prescribed with different hypoglycemia drug classes, where Ratio=Number/TotalNumber (and the TotalNumber is 620,633). Insulin is the most frequently used hypoglycemia drug class, followed by AGIs and Biguanides, while DPP-4 is rarely used in our cohort.

**Learning Framework**

As its name implies, "Deep Diabetologist" is a cognitive decision advisor who has deep insights by learning from longitudinal EHR data to provide personalized hypoglycemia medication predictions for diabetic patients. In this respect, we need to clarify (1) what to learn, and (2) how to learn.

In real life, for diabetic patients whose blood glucose has not been well controlled, two or more hypoglycemia drug classes are often combined together as medications. Recalling to Figure 1, on 2015-08-24, that patient was prescribed with biguanides plus sulfonylureas. By counting the number of distinct drug combinations, we get the number of 85 (theoretically, the combination number could be $2^7-1 = 127$, but not all drug combinations are clinically meaningful).

*Table 4. Classification problem for predicting medications*

| Problem | Binary Classification | | Multiclass Classification | |
|---|---|---|---|---|
| | $BC_{DC}$ | $BC_{DCC}$ | **$MC_{DC}$** | **$MC_{DCC}$** |
| # of classifiers | 7 | 85 | **1** | **1** |
| # of classes | 1 | 1 | **7** | **85** |
| Multiple label? | No | No | **Yes** | **No** |
| Evaluation | AUC | Accuracy | **AUC** | **Accuracy** |

Now, the problem comes – which type of classification problem should we formulate? Table 4 lists the different classification problems for predicting medications, where # of classifiers means the number of classifiers to be trained, and # of classes means the number of classes to be labeled. In this paper, we will present our experimental results for the multiclass classification problem (in bold): one classifier for 7 drug classes (as denoted as $MC_{DC}$) and the other classifier for 85 drug combination classes (as denoted as $MC_{DCC}$). In real clinical settings, some junior physicians, such as general practitioners, are more interested in $MC_{DCC}$ which predicted results – after the final confirmation – could be directly fed into the computerized physician order entry (CPOE). However, for some senior physicians, such as specialists, might be more interested in $MC_{DC}$, which predicted results need the final decision for combination and then to be fed into CPOE. Therefore, we keep experiments for both $MC_{DC}$ and $MC_{DCC}$, while leaving the binary classification problems for future work.

*RNN medication prediction model*

We start by defining a recurrent neural network medication prediction model (namely, RNN_MPM), referring to the well-established recurrent neural network language model [11].

The RNN_MPM is a probabilistic generative model, with parameters $\theta$ which decomposes the probability over medical visits $v_1, ..., v_N$:

$$P_\theta(v_1, ..., v_N) = \prod_{n=1}^{N} P_\theta(v_n|v_1, ..., v_{n-1})$$

For the one-label problem $MC_{DCC}$, the parameterized approximation of the output distribution uses an RNN with a softmax layer on the top, where $dcc$ (or $dcc'$) is one of the 85 drug combination classes:

$$P_\theta(v_{n+1} = dcc|v_1, ..., v_n) = \exp(g(h_n, dcc)) \Big/ \sum_{dcc'} \exp(g(h_n, dcc'))$$

For the multi-label problem $MC_{DC}$, the parameterized approximation of the output distribution uses an RNN with multiple sigmoid layers on the top, where $dc$ is one of the 7 drug classes:

$$P_\theta(v_{n+1} = dc|v_1, ..., v_n) = 1 \Big/ 1 + \exp(-g(h_n, dc))$$

For both $MC_{DCC}$ and $MC_{DC}$, $f$ is the hidden state update function, which we will assume is either an LSTM gating unit [12] or a GRU gating unit [13]. In this paper, we will use the LSTM gating unit for experiments.

$$h_n = f(h_{n-1}, v_n)$$

The function $g$ is the dot product of $h_n$ and $dc$ (or $dcc$), where $dc$ is a multi-hot vector of the 7 drug classes, and $dcc$ is a one-hot vector of the 85 drug combination classes.

Figure 3 shows a basic RNN medication prediction model in our scenario, where the output $Y$ is, either a multi-hot vector of the 7 drug classes for $MC_{DC}$, or a one-hot vector of the 85 drug combination classes for $MC_{DCC}$, and the subscript $k$ means the $k^{th}$ time step. For the input, $X$ is a vector of multivariate values, including real values (for physical examinations and lab tests) and binary values (for diagnosis codes and previously prescribed drug classes).

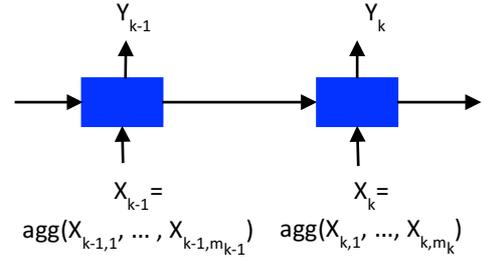

*Figure 3. A basic RNN medication prediction model.*

Here, we use an aggregation function, $X_k=agg(X_{k,1}, ..., X_{k,mk})$. In fact, the $k^{th}$ time step has an observed medication, and from the $(k-1)^{th}$ time step to the $k^{th}$ time step, there is no medication observed, but there could be other observations like diagnoses, physical examinations and lab tests. Recalling to Figure 1, the $(k-1)^{th}$ time step was 2015-06-15 using biguanides, the $k^{th}$ time step was 2015-08-24 using biguanides plus sulfonylureas, and on 2015-07-08, the patient had blood pressure measurements with diagnosis of hypertension but without any hypoglycemia medication. Therefore, we collect all observations at every time step, and $X_{k,1}, ..., X_{k,mk}$ are meant to be $m_k$ time steps sandwiched between the $(k-1)^{th}$ medication and the $k^{th}$ medication. These vector values will be aggregated by a function (e.g. average, maximum, count), to be input at the $k^{th}$ time step.

*HRNN medication prediction model*

Following up, we naturally wonder whether we can leverage a hierarchical RNN framework for modeling the sandwiched time steps, as shown in Figure 4. For the upper RNN layer, its output definition is the same as in Figure 3, while its input is the output from a lower RNN layer. For the lower RNN layer, its input is now the vector at each sandwiched time steps, $X_{k,1}, ..., X_{k,mk}$, whose output will be input to the upper RNN layer.

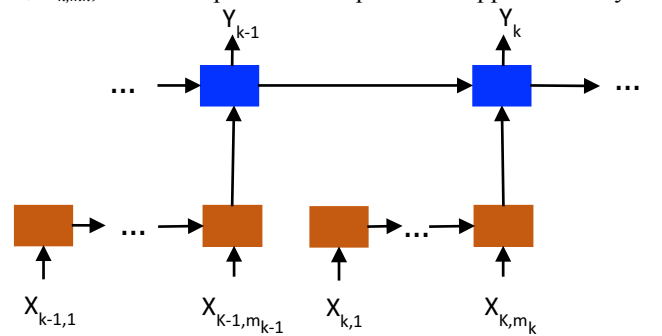

*Figure 4. A hierarchical RNN medication prediction model.*

Figure 5 shows a more complicated model, where the input of the upper RNN layer is now a merge of multiple outputs from lower RNN layers. This somehow matches our intuition, if we regard the green RNN layer as learning clinical measurements, and the yellow RNN layer as learning diagnoses, then such a hierarchical RNN model is to predict medications based on an abstract representation by learning from multi-modality sequences of clinical measurements and diagnoses.

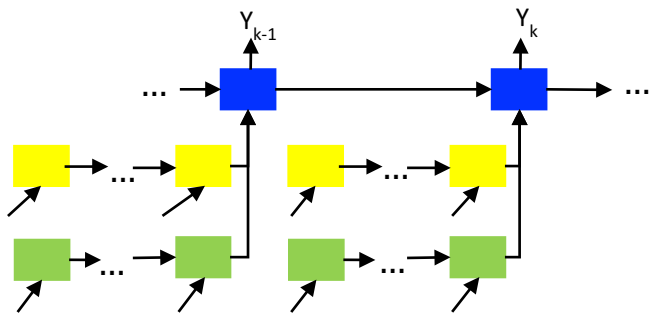

*Figure 5. A hierarchical RNN model with multiple layers.*

Here, we claim, although HRNN has been proposed and implemented in domains such as natural language processing and computer vision [8][9][10], the HRNN application for predicting medications is firstly investigated in this paper, which deeply explores the heterogeneity of EHR data in multivariate time series.

## Results

For experiments, we use a cohort of 21,796 patients (as introduced in the section of Data Preprocessing) from an EHR repository of a level 2 city in China. The cases to be studied are outpatient hypoglycemic prescriptions after the firstly observed diabetes diagnosis, where a case is counted per day per patient, and the number of cases is 620,633. All models are trained on 80% of the cohort and validated on 10%, remaining 10% for testing. Table 5 is an overview of the experimental dataset.

*Table 5. Overview of dataset*

|  | Number of patients | Number of cases |
|---|---|---|
| Training | 17,436 | 496,252 |
| Testing | 2,180 | 62,497 |
| Validation | 2,180 | 61,884 |
| Total | 21,796 | 620,633 |

Our experiments are not meant to verify the RNN power, which has been well shown up in related works, e.g. [5][6][7]. Instead, we are more interested in a practical model which could be applied, in real clinical settings, to provide decision support. So, we do experiments using different (basic/hierarchical) variants of RNN models to address the multi-class problems for predicting either multiple labels of the 7 drug classes (a.k.a., $MC_{DC}$) or one label of the 85 drug combination classes (a.k.a., $MC_{DCC}$).

In this paper, LSTM is applied to implement the RNN, on a machine configured with GPU using Theano backend. We train each LSTM for 20 epochs (actually, we have observed the convergent results after 10 epochs), and to avoid overfitting, we stack a dropout (of 0.5) layer before the last Dense layer of the output vector. Besides, to address the variable-length inputs to RNN, we set a maximal length of 20 for input (if an instance has length less than 20, then we will mask the rest as 0), and a masking layer with mask value of 0. All hidden layers are set up with 64 cells, and the batch size is 32.

**Experiment 1: with previous medications**

As mentioned above, in real life, the prescriptions are often the same as before. Therefore, for the first experiment, we will compare our (H)RNN models with the baseline, assuming previous medications are available for the input. Then, we will conduct a second experiment, when previous medications are absent for the input.

In Experiment 1, we have 4 predictors: Prev., LR, RNN and HRNN. The Prev. uses the previous medication for the current prediction. LR (Logistic Regression), as usual, is the best candidate for a baseline classifier, with input of 481=350+124+7 variables, where 350 is the number of ICD codes, 124 is the number of clinical measurements, and 7 is the number of previously used drug classes. RNN and HRNN are defined as above, where RNN has the same 481 variables for the input at every time step (referring to Figure 3), and HRNN has an upper layer of 7 drug classes with merge of the outputs from two lower layers (referring to Figure 5), where one layer (in yellow) inputs a vector of 350 ICD codes and the other layer (in green) inputs a vector of 124 clinical measurements.

Table 6 and Table 7 are results for the multi-label problem $MC_{DC}$ and the one-label problem $MC_{DCC}$, respectively. In Table 6, AUC is the area under the curve, and we report the AUC scores for 7 drug classes as well as the average. In Table 7, accuracy = #Hit/#Sample, where #Hit is the number of cases in which the predicted label equals to the targeted label, and #Sample is the number of samples. Moreover, we separately report the different accuracy results for those head cases, tail cases, and all cases – the head case means the first medication of a patient's sequence, and the tail case means the last medication of a patient's sequence, while the average is calculated for all cases of a patient's sequence. The head accuracy of Prev. is not available, since there is no previously prescribed medication to count in for the head case.

*Table 6. AUC results for $MC_{DC}$ in Experiment 1*

|  | Prev. | LR | RNN | HRNN |
|---|---|---|---|---|
| Biguanides | 0.8630 | 0.8928 | 0.9252 | **0.9259** |
| Sulfonylureas | 0.8798 | 0.9210 | 0.9428 | **0.9433** |
| Glinide | 0.8426 | 0.8676 | 0.9326 | **0.9339** |
| TZDs | 0.8245 | 0.8481 | 0.9169 | **0.9229** |
| AGIs | 0.8541 | 0.8305 | 0.9104 | **0.9133** |
| DPP-4 | 0.8823 | 0.8761 | **0.9058** | 0.9051 |
| Insulin | 0.9164 | 0.8905 | 0.9422 | **0.9435** |
| Average | 0.8661 | 0.8752 | 0.9251 | **0.9268** |

*Table 7. Accuracy results for $MC_{DCC}$ in Experiment 1*

|  | Prev. | LR | RNN | HRNN |
|---|---|---|---|---|
| Head | -- | 0.2963 | 0.3417 | **0.3436** |
| Tail | 0.6772 | 0.6986 | 0.7060 | **0.7067** |
| Average | 0.6456 | 0.6693 | 0.6733 | **0.6745** |

Remarkably, results of Prev. are quite high – by average, the accuracy is 0.6456 and the AUC is 0.8661. The performance of LR is slightly improved, by comparison of Prev. Our RNN and HRNN both outperform Prev. and LR. From HRNN to RNN, the performance is increased (as shown in bold), except the AUC score of DPP-4. We argue that's because of the very limited number of DPP-4 cases (referring to Table 3).

**Experiment 2: without previous medications**

Next, we have only 3 predictors: LR, RNN and HRNN, since that we assume previous medications are not available for the input – This experiment is meant to remove the "big plays" of previous medications.

Similar to Experiment 1, LR is still the baseline classifier, with input of 474=350+124 variables, where 350 is the number of ICD codes, and 124 is the number of clinical measurements.

RNN has the same 474 variables for the input at every time step (referring to Figure 3), and HRNN has an upper layer with no raw data input from those 7 drug classes, but only a merge of the outputs from two lower layers (referring to Figure 5).

Results for $MC_{DC}$ and $MC_{DCC}$ are reported in Table 8 and Table 9, respectively. Both RNN and HRNN outperform the baseline LR. The performance of HRNN is slightly higher than that of RNN, and again, the AUC score of DPP-4 is the exception.

*Table 8. AUC results for $MC_{DC}$ in Experiment 2*

|  | LR | RNN | HRNN |
| --- | --- | --- | --- |
| Biguanides | 0.6254 | 0.7088 | **0.7091** |
| Sulfonylureas | 0.6095 | 0.6328 | **0.6421** |
| Glinide | 0.5774 | 0.5967 | **0.6115** |
| TZDs | 0.5184 | 0.6013 | **0.6058** |
| AGIs | 0.5766 | 0.5548 | **0.5698** |
| DPP-4 | **0.7703** | 0.7269 | 0.7043 |
| Insulin | 0.5855 | 0.6475 | **0.6520** |
| Average | 0.6090 | 0.6384 | **0.6421** |

*Table 9. Accuracy results for $MC_{DCC}$ in Experiment 2*

|  | LR | RNN | HRNN |
| --- | --- | --- | --- |
| Head | 0.2932 | 0.3404 | **0.3408** |
| Tail | 0.3454 | 0.3477 | **0.3495** |
| Average | 0.3484 | 0.3480 | **0.3509** |

## Discussion

Deep learning techniques have been successfully applied to various fields, such as computer vision, natural language processing, speech and image recognition. In healthcare domain, the disease risk model was deeply trained in [7], clinical phenotypes were incrementally discovered in [14], and [15] proposed a general-purpose patient representation from EHR data. Our work follows this trend, and our experimental results demonstrate the use of (hierarchical) recurrent neural networks to model sequential events in EHR data would provide deep evidence for treatment recommendation.

However, we have to admit our limitations. First, the improvement from HRNN to RNN is not exciting, and we owe that to the serious shortage of clinical measurements in our EHR repository. For our proposed HRNN models (referring to Figure 5), we have a lower RNN layer with input of sequential clinical measurements, and the scarcity of clinical measurements undoubtedly makes our HRNN models less effective. Second, in HRNN implementation, we simply concatenate the hidden states from lower RNN layers, and more advanced merge functions should be considered in future. Third, from Experiment 1 to Experiment 2, the total performance is decreased, which again proves the strength of prescribing as previously. An interesting ongoing work is to use our current (H)RNN models to implement a binary classifier to predict the prescription change – that is, the outputs of Figure 3,4,5 become binary values (1 for prescription being changed, 0 for prescription being unchanged), while still using the inputs as defined above.

## Conclusions

"Deep Diabetologist" is named for our work, and the rationale behind this naming is to provide clinical decision support, by deep learning from the longitudinal EHR data. In this paper, we propose basic and hierarchical RNN medication prediction models, and by experiments, we demonstrate the improved performance of predicting hypoglycemia medications for diabetic patients. In particular, we pave the way for using the hierarchical RNN framework for EHR data modeling.

Admittedly, our current work of "Deep Diabetologist" is a naïve learner, without any ability to distinguish good and bad clinical outcomes. Instead, the goal of clinical decision support is to improve outcome, and how to train our "Deep Diabetologist" towards good outcomes is still a long way to go.

**Address for correspondence**

Jing Mei, PhD. IBM Research China, Building 19, Zhongguancun Software Park, Haidian District, Beijing, China, 100193.
Email: meijing@cn.ibm.com